\documentclass[11pt,a4paper]{article}
\usepackage[hyperref]{acl2022}
\usepackage{xcolor}
\usepackage{times}
\usepackage{latexsym}
\usepackage{multirow}
\usepackage{array}
\usepackage{fixltx2e}
\usepackage[ruled]{algorithm2e}
\usepackage{amsmath,amssymb,amsfonts,enumitem}
\newcolumntype{P}[1]{>{\centering\arraybackslash}p{#1}}

\usepackage{microtype}
\usepackage{graphicx}

\aclfinalcopy 
\def\aclpaperid{163} 

\newcommand{\system}{{\textsc{CompactIE}}}

\title{CompactIE: Compact Facts in Open Information Extraction}

\author{Farima Fatahi Bayat\\
  University of Michigan \\
  \texttt{farimaf@umich.edu} \\\And
  Nikita Bhutani \\
  Megagon Labs \\
  \texttt{nikita@megagon.ai} \\\And
  H. V. Jagadish \\
  University of Michigan \\
  \texttt{jag@umich.edu} \\}

\date{}

\begin{document}
\maketitle
\begin{abstract}

A major drawback of modern neural OpenIE systems and benchmarks is that they prioritize high coverage of information in extractions over compactness of their constituents.
This severely limits the usefulness of OpenIE extractions in many downstream tasks.
The utility of extractions can be improved if extractions are compact and share constituents. To this end, we study the problem of identifying compact extractions with neural-based methods. We propose \system{}, an OpenIE system that uses a novel pipelined approach to produce compact extractions with overlapping constituents. It first detects constituents of the extractions and then links them to build extractions. We train our system on compact extractions obtained by processing existing
benchmarks. Our experiments on CaRB and Wire57 datasets indicate that \system{} finds 1.5x-2x more compact extractions than previous systems, with high precision, establishing a new state-of-the-art performance in OpenIE.

\end{abstract}

\section{Introduction}

{A popular domain-agnostic paradigm to structure the raw text is open information extraction (OpenIE) \cite{textRunner}. Not relying on any pre-defined schema, OpenIE systems typically extract information as (subject; relation; object) triples. The extracted information is then used in several downstream applications, including answering questions~\cite{DBLP:journals/corr/KhotSC17}, summarizing documents~\cite{hao2018structured, ji2013open}, and populating knowledge bases~\cite{DBLP:journals/corr/abs-1910-08435}}.

\begin{figure}
\centering
\includegraphics[width=0.48\textwidth]{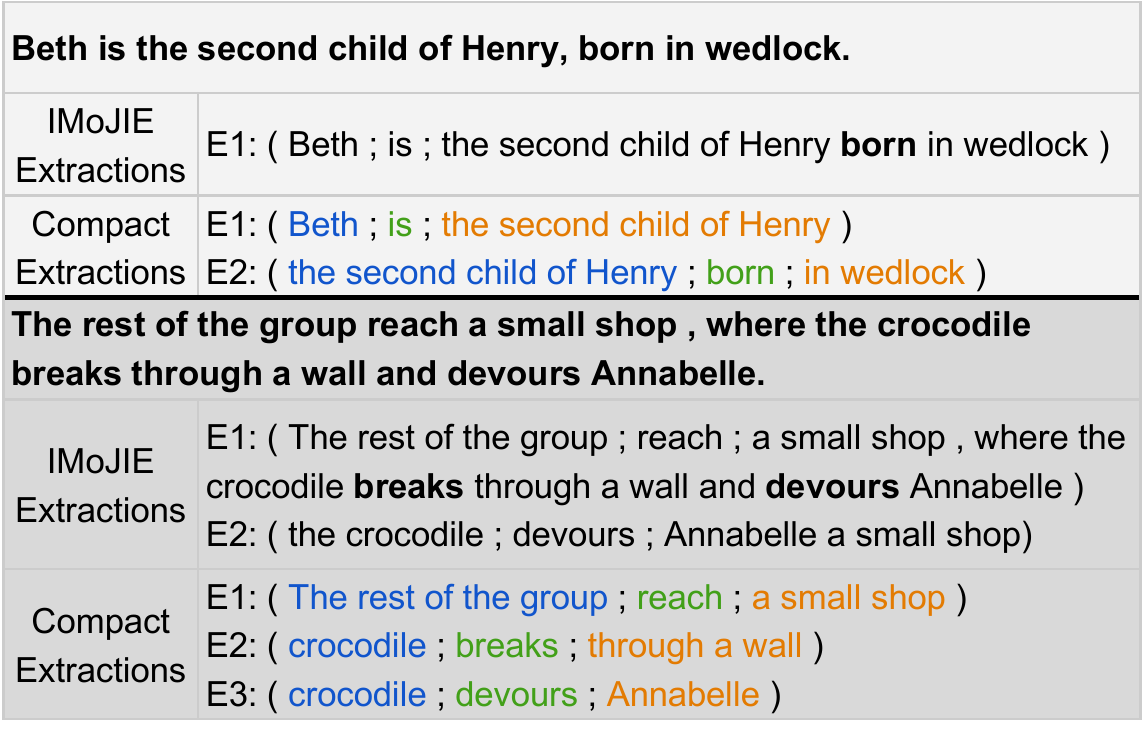}
\caption{Example sentences with non-compact triples from IMoJIE vs. compact triples from our benchmark. Compact triples can share constituents. Constituents for subjects, relations and objects are indicated in \textcolor{blue}{blue}, \textcolor{teal}{green} and \textcolor{orange}{orange}, respectively.}
\label{fig:exp1}
\vspace{-1.4em}
\end{figure}

{Despite much progress, state-of-the-art neural OpenIE systems focus on covering more information from the input sentence often at the cost of utility and compactness of the extracted triples. The extracted triples have long, over-specific {\it constituents} (i.e. the relation and its arguments).
Figure \ref{fig:exp1} illustrates such example triples produced by a popular OpenIE system, IMoJIE \cite{imojie}.
As shown, the knowledge that \emph{the second child of Henry} was born \emph{in wedlock} is embedded in a long argument. This can be problematic for downstream applications, especially knowledge base population~\cite{gashteovski2020aligning,stanovsky2015open} that derive power from merging multiple pieces of information extracted about the same entity. In contrast, the compact extractions are more pliable for tasks such as identifying similar facts and merging facts that share constituents. For example, compact extractions in Figure \ref{fig:exp1} can be merged to derive that \emph{Beth} is born \emph{in wedlock}.}

{Although some prior work \cite{clausie,minie,nestie} has explored the compactness of OpenIE triples, these systems are rule-based and have been superseded by end-to-end neural OpenIE systems.}
In this work, we study the problem of identifying compact extractions with neural-based methods. Inspired by \citep{clausie}, we define an extracted triple to be {\it compact} if it does not contain information that can be independently represented in another triple. To further improve the suitability of compact triples for knowledge base population, we require compact triples extracted from a sentence to have overlapping constituents.

Existing neural systems adopt a {sequence labeling} \cite{openie6, wang2021unire,multi2OIE} or a {sequence generation} \cite{imojie} approach to identify triples and their constituents, typically all at once, or {through a pipeline that first identifies the relations and then their corresponding arguments. None of these methods guarantee that the extracted triples will be compact and share constituents.}

We propose a novel pipeline system for finding compact triples that share their constituents. We call our OpenIE system, \system{}. {To encourage the constituents to be shared across triples, \system{} first extracts the constituents using a  {\it Constituent Extraction} model and then links them using a {\it Constituent Linking} model to obtain triples.}

We adapt a table filling method \cite{wang2021unire} with a new schema for identifying both constituent boundaries and their roles (i.e., subject or object). This allows the constituent extraction model to capture interactions among constituents and minimize ambiguities in boundary detection. {For the task of constituent linking, we train a model that builds on contextual representations specific to a given pair of constituents and predicts their relation type.}
Such a two-step approach enables us to optimize the models for each sub-task with different objectives and also promote the constituent reuse across triples.

Existing neural OpenIE systems are trained on benchmarks that combine extractions from multiple OpenIE systems .
However, no such large-scale benchmark exists for compact triples. We develop a new benchmark using a subset of sentences in the OpenIE2016 benchmark \cite{openie4}. Specifically, we develop a data processing algorithm that targets extraction from individual clauses in a sentence. Given an input sentence, it identifies clauses and then uses OpenIE systems such as IMoJIE over the clauses to find compact triples. We train \system{} on the new benchmark.

Our experiments on a fine-grained benchmark, Wire57, show that \system{} outperforms existing non-neural and neural systems by 5.8 F1 pts and 7.1 F1 pts,
respectively. Manual evaluation over a coarse-grained benchmark, CaRB, indicates that \system{} produces 1.5x-2x more compact extractions than existing systems with comparable precision, establishing a new state-of-the-art for the OpenIE task\footnote{Source code, benchmark dataset, and related resources are available at \url{https://github.com/FarimaFatahi/CompactIE}}.

\section{Background and Preliminaries}
\label{'task'}
Given a sentence $s = w_1 w_2 ... w_n$, an OpenIE system generates triples of the form \emph{(subject; relation; object)}, where \emph{subject}, \emph{relation} and \emph{object} are the constituents of a triple. 

\subsection{Extracting Compact Triples}

A recent study~\cite{gashteovski2020aligning} shows that triples from modern neural OpenIE systems are difficult to align to knowledge bases such as DBpedia. Less than 77\% of triples from neural OpenIE systems had the same arguments as DBpedia facts. In contrast, the corresponding alignment ratio for some of the non-neural OpenIE systems was as high as 98\%. They attribute this behavior to the specificity of the triples. A compact triple, which does not contain complete clauses as part of a constituent or contain additional information, is easier to align to DBpedia. {Our goal is to leverage neural-based methods to extract compact triples.}

\subsection{System Architecture}

We focus on extraction from individual clauses within a sentence, where each clause includes a subject, a verb, optionally a direct object, and a compliment. Since extractions from different clauses share information, we split the OpenIE task into two sub-tasks: {\it constituent extraction} and {\it constituent linking}.

The task of constituent extraction is to find a set of constituents such that each constituent $c$ is a contiguous span of words $c.span = \{(w_i, w_j)\}$ and has a pre-defined type $c.type \in Y_c$ where $Y_c = \{Argument, Predicate\}$. The constituent that takes the \emph{relation} role in a triple has $c.type = Predicate$, and \emph{subject} and \emph{object} constituents have $c.type=Argument$.
This schema simplifies the task and provides more information to the constituent linking model.
 
The task of constituent linking is to connect a given set of $Predicate$ constituents $\{p_1$, \dots $p_m\}$ and $Argument$ constituents $\{a_1$, \dots $a_n\}$ to obtain triples. We formulate this as a relation classification task where the set of relations is $Y_r = \{Subject, Object\}$. The model predicts relations $r$ between each $p_x$ and $\{a_1$, \dots $a_n\}$ such that:
\vspace{-0.5em}
\[\exists (i, j): r(a_i, p_x) = Subject \hspace{2pt} , \hspace{2pt} r(p_x, a_j) = Object\]
to construct triple $(a_i$; $p_x$; $a_j)$.

\begin{figure}[ht!]
\includegraphics[width=0.48\textwidth]{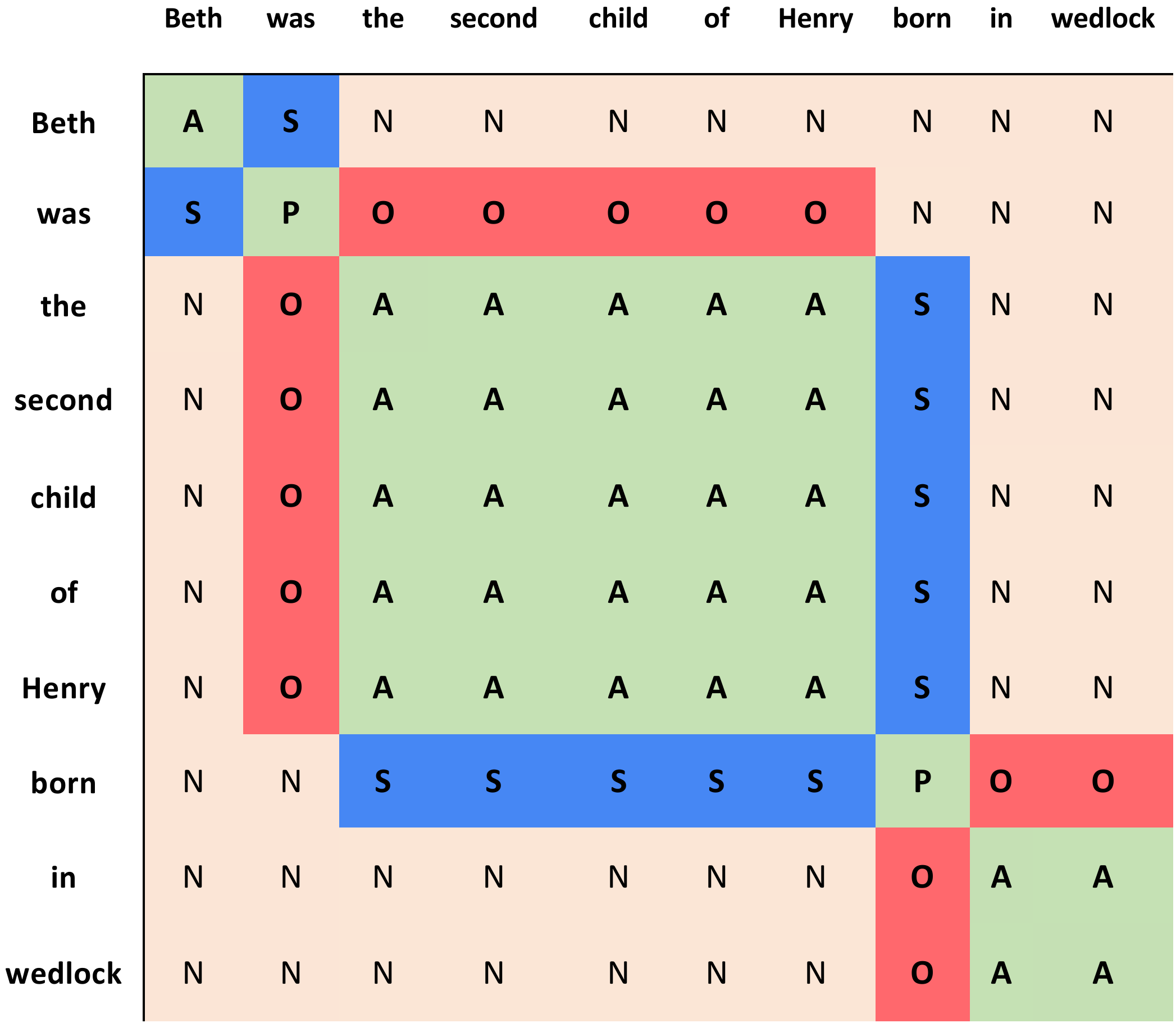}
\caption{Table filling based on the relation between each pair of words in the sentence. Argument (\textbf{A}) and Predicate (\textbf{P}) are constituent types. Subject (\textbf{S}) and Object (\textbf{O}) declare the relation between two constituents (\textbf{N} stands for no relation).}
\label{fig:constitunet_table}
\vspace{-1em}
\end{figure}

\begin{figure*}[t]
\centering
\includegraphics[width=\textwidth]{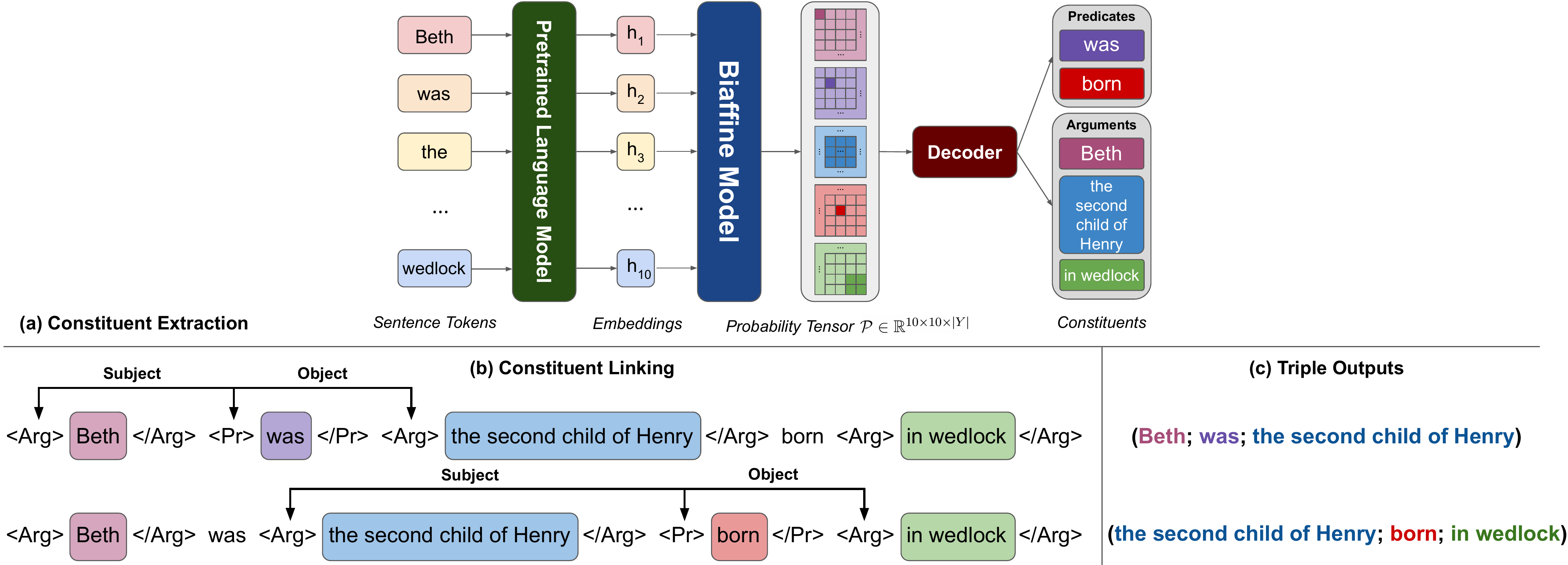}
\caption{
Overview of system architecture. Given the sentence: ``Beth was the second child of Henry, born in wedlock.", the Constituent Extraction model identifies the span and type of constituents (top-right). Next, the Constituent Linking model (b) searches for $Argument$s of each $Predicate$ constituent independently. Thus, for each of the two extracted $Predicate$s, it modifies the input sentence by inserting typed constituent markers (\(<\)Arg\(>\), \(<\)/Arg\(>\) to specify the start and end of arguments and \(<\)Pr\(>\), \(<\)/Pr\(>\) for predicates). Finally, the modified sentence is fed into a classifier to find $Subject$ and $Object$ of each $Predicate$ and form triples (c).}
\label{fig:SystemArchitecture}
\vspace{-1em}
\end{figure*}

\section{Approach}
{In this section we describe our pipeline system, \system{}. We first detail the constituent extraction model, its training constraints, and the decoding algorithm in Section \ref{constituent_extraction}. Then, we describe the constituent linking model in Section \ref{constituent_linking}. Figure \ref{fig:SystemArchitecture} shows an overview of \system{} architecture.}

\subsection{Constituent Extraction Model}
\label{constituent_extraction}

The constituent extraction model aims to find constituent spans and their types in a sentence. 
{Following recent progress in entity-relation extraction~\cite{wang2021unire}, we model this as a table filling problem. However, we design a new table schema for the constituent extraction task. 
Figure~\ref{fig:constitunet_table} shows an example schema. A sentence $s$ with $|s|$ tokens corresponds to a table $T^{|s| \times |s|}$ such that each cell is labeled based on the relation between the pair of words. For each constituent, corresponding cells are labeled {with $y_c \in \{Argument, Predicate\}$}. For relations between different constituents, corresponding cells are labeled {with $y_r \in \{Subject, Object\}$}. The cells with no relations are labeled $None$. Graphically, constituents are squares on the diagonal, and relations are rectangles off the diagonal. 

\subsubsection{Table Filling Model}
\label{'sec3.1.1'}

Given the tabular formulation, the constituent extractor performs two tasks: a) fill the table by predicting labels for each word pair, b) extract the constituents given the label probabilities. Following~\citep{wang2021unire}, we adopt a biaffine attention mechanism, described next, to learn interactions between word pairs when filling the table.}

\noindent
Given the input sentence $s$, we first obtain contextual representation $\mathbf{h_i}$ for each word using a pre-trained language model (e.g. BERT~\cite{DBLP:journals/corr/abs-1810-04805}). We then employ two MLPs to identify the head and tail role of the word given its vector representation $h_i$.

\vspace{-1.75em}

\[h_i^{head} = MLP_{head}(h_i), h_i^{tail} = MLP_{tail}(h_i)\]

\vspace{-0.5em}

\noindent
Next, using the biaffine scoring function, we calculate the scoring vector of each pair of words (e.g. $w_i, w_j$) as follows:

\vspace{-1.75em}

\[t_{i,j} = (h_i^{head})^TU^{(1)}h_j^{tail} + (h_i^{head} \oplus h_j^{tail})^TU^{(2)} + b\]

\vspace{-0.5em}

\noindent
where $U^{(1)}, U^{(2)}$ are weight parameters, $b$ is the bias term and $\oplus$ denotes concatenation. {Then, we feed the score vector $t_i,j$ into a softmax function to calculate the probability distribution of the corresponding labels $l \in Y$, where $Y = Y_c \cup Y_r \cup {None}$.}

\vspace{-1em}

\[P(y_{i,j}| s) = Softmax(t_{i,j})\]

\noindent
Finally, we train the 2D table to minimize the following training objective:
\vspace{-1em}
\[L_{entry} = -\frac{1}{|s|^2} \sum_{i=1}^{|s|}\sum_{j=1}^{|s|} \log(P(y_{i,j} = Y_{i,j} | s))\]

\vspace{-0.5 em}

\noindent
where \hspace{-1pt}$Y_{i,j}$\hspace{-0.1pt} is the gold label for cell $(i,j)$\hspace{-1pt} in the table. 

\subsubsection{Training Constraints}
\label{'sec3.1.2'}

{\citep{wang2021unire} shows that structural constraints imposed on the table during training can significantly enhance the model. We adopt their {\it symmetry} and {\it implication} constraints. However, we observed that these alone are not sufficient if certain labels are preferred over others. For example, all triples must have a subject, but some may not have an object. We propose a new {\it triple} constraint to further enhance our model. In this section, we describe the three constraints in detail. We also introduce $\mathcal{P} \in R^{|s|\times|s|\times|Y|}$ that denotes the stack of $P(y_{i,j} |s)$ for all word pairs in sentence $s$.}

\noindent
\textbf{Symmetry}: This constraint ensures that the table is symmetric i.e. the squares are symmetric about the diagonal. As shown in Figure \ref{fig:constitunet_table}, this ensures the label assigned to the (second, Henry) cell is the same as the cell (Henry, second). Given matrix $\mathcal{P}$, We formulate this constraint as symmetrical loss:
\vspace{-1em}
\[L_{sym} = -\frac{1}{|s|^2} \sum_{i=1}^{|s|}\sum_{j=1}^{|s|}\sum_{t \in Y_r \cup Y_c} |\mathcal{P}_{i,j,t} - \mathcal{P}_{j,i,t}|\]
\vspace{-0.5em}

\noindent
\textbf{Implication}: This constraint implies that no relation would appear unless its constituents are present in the table. This is imposed on $\mathcal{P}$: for each word in the diagonal, maximum possibility over the constituent type space $Y_c = \{Argument, Predicate\}$ is not lower than the maximum possibility for other words in the same row or column over the relation type space $Y_r = \{Subject, Object\}$.
\vspace{-0.3em}
\begin{equation*} 
L_{imp} = \frac{-1}{|s|} \sum_{i=1}^{|s|} \left[ \max_{t \in Y_r}(\mathcal{P}_{i,:,t}, \mathcal{P}_{:,i,t}) - \max_{t \in Y_c}(\mathcal{P}_{i,i,t})\right]_*\footnote{$[u]_* = max(u, 0)$ is the hinge loss.}
\end{equation*} 
\vspace{-0.5em}

\noindent
\textbf{Triple Constraint}: 
{This constraint enables the model to increase the likelihood of certain roles (e.g. \emph{\textbf{S}ubject}) over the others (e.g. \emph{\textbf{O}bject}) to ensure the triples are valid. We enforce this constraint on $\mathcal{P}$: For each column or row corresponding to a \emph{Predicate} constituent, the maximum possibility of off-diagonal words over $Subject$ type is not lower than the maximum possibility of off-diagonal words over $Object$ type. We formulate this constraint as triple loss. }

\vspace{-0.5em}
\begin{equation*}
\begin{split}
    L_{triple} &= \frac{-1}{2|ps|}\sum_{i\in{ps}} \biggr[ \{max(\mathcal{P}_{i,:,\boldsymbol{O}}) - max(\mathcal{P}_{i,:,\boldsymbol{S}}\} \\ 
    & + \{max(\mathcal{P}_{:,i,\boldsymbol{O}}) - max(\mathcal{P}_{:,i,\boldsymbol{S}})\} \biggr]
\end{split}
\end{equation*}
\vspace{-0.2em}

\noindent
where $ps$ is union of $Predicate$ spans in sentence.

Finally, we jointly optimize four objectives in training: $ L_{entry}+ L_{sym} + L_{imp} + L_{triple}$

\subsubsection{Decoding}
\label{'sec3.1.3'}

Given the label probability tensor $\mathcal{P}$, we need to decode the constituents in the testing phase. We follow a 2-step decoding procedure that finds spans of constituents first and then assigns a label to each span. The decoder first calculates the distance between adjacent rows and columns of the table to find constituents' boundaries. Next, it assigns a type to each span and filters out any $None$ constituents before passing the output to the linking model. The upper part of Figure \ref{fig:SystemArchitecture} shows the output of the decoder, which extracts two constituents (``was", ``born") of type $Predicate$ and three constituents (``Beth", ``the second child of Henry", ``in wedlock") of type $Argument$. We provide a detailed description of the decoding algorithm in Appendix \ref{sec:2stepDecoding}.

\subsection{Constituent Linking Model}
\label{constituent_linking}

The constituent linking model aims to take a $Predicate$ constituent and a set of $Argument$ constituents as input and predict a relation label $Y_r = \{Subject, Object, None\}$. This procedure is repeated for each predicate constituent in the sentence. We formulate this as a relation classification task where the model classifies relation labels of given constituent pairs based on context. 

Following prior work~\cite{zhang2019ernie,DBLP:journals/corr/abs-2010-12812}, we modify the token sequence of input sentence by adding marker tokens \(<\)Pr\(>\), \(<\)/Pr\(>\), \(<\)Arg\(>\), \(<\)/Arg\(>\) to highlight the constituent spans and their types. The markers help the linking model combine context information and constituent information for relation classification. As shown in Figure \ref{fig:SystemArchitecture}.a, two types of constituents are extracted from the input sentence. For each constituent of type $Predicate$, we modify the input sentence by highlighting the location of the $Predicate$ and all $Argument$ constituents. Then, we feed this processed sentence to a pre-trained encoder (BERT).

{Next, we concatenate the output representation of the start position of predicate $p$ with the output representation of the start position of argument $a_i$: }

\vspace{-0.75em}

\[X_r(p, a_i) = h_{start(p)} \oplus h_{start(a_i)}\]

\noindent
Finally, we feed the concatenated representation into a multi-layer perceptron (MLP) to predict the probability distribution of the relation type $ r \in Y_r \cup {None}$: 

\vspace{-1em}

\[P(r | p, a_i) = MLP(X_r)\]

\vspace{-0.5em}

\section{Benchmark Creation}
\label{benchmark}

To train the constituent extraction and constituent linking models for extracting compact triples, we need a benchmark of compact triples.
Existing OpenIE benchmark
\footnote{\url{https://github.com/dair-iitd/imojie/tree/master/benchmark}} 
is created by combining extractions from multiple existing OpenIE systems. 
Although widely adopted, we observed that it includes over-specific and sometimes incorrect extractions from previous systems. This encouraged us to design a data processing algorithm that can extract compact triples from scratch. {Inspired by rule-based OpenIE system \cite{clausie}, we find compact triples by extracting the following clauses within a sentence:}

\noindent
{\textbf{Main Clauses} are independent clauses that express a complete concept.} 

\noindent
\textbf{Complement Clauses} are subordinate clauses that serve to complete the meaning of a verb or noun in the sentence. 

\noindent
\textbf{Coordinate Clauses }are independent clauses joined to the main clause using coordinating conjunctions such as {\em and}, {\em or}, {\em but}, etc.

\noindent
We identify clauses within a sentence using its dependency graph. We first build a sentence tree such that the root is the head of the main clauses and the first-level children are clauses modifying the root word. We then perform a postfix traversal of the tree until we find a sub-tree with no clausal children. At this point, we run a standard OpenIE system, IMoJIE~\cite{imojie}, over the clause corresponding to the sub-tree to obtain triples. We then backtrack and extract triples for other clausal children and lastly the parent. We provide pseudo-code of algorithm in Appendix \ref{sec:appendix}.

\noindent
We run our algorithm on each multi-clause sentence in the OpenIE2016 benchmark and obtain a new benchmark tailored for extracting compact triples. Figure \ref{fig:exp1} shows example sentences and compact triples from this benchmark.

\section{Experimental Setup}
\label{sec:experimental_setup}

\noindent
\textbf{Training Dataset:}
We train \system{} using the benchmark described in Section \ref{benchmark}. Table \ref{datasets_statistics} compares the statistics of our new benchmark and bootstrapped OpenIE2016 benchmark. As shown, our benchmark has 1.25 times more extractions per sentence than OpenIE2016 and its constituents are more compact. We use about 1\% of sentences for validation and the remaining for training.

\begin{table}
\begin{center}
\resizebox{\linewidth}{!}{
\begin{tabular}{ccccc} 
\hline
& \multicolumn{3}{c}{\textbf{Our Benchmark}} & {\normalsize \textbf{OIE2016}} \\
\cline{2-4}
& { \textbf{Total}} & {Train} & {Valid} &  \\
\hline
{{\# Sentence}} & 54.9k & 54.5k & 500 & 92.7k \\
{{\# Triples}} & 121.8k & 120.6k & 1155 & 190.6k \\
\hline
Avg. \# triples per sent. & \textbf{3.165} & - & - & 2.542 \\
Avg. constituent length & \textbf{4.587} & - & - & 7.893 \\
\end{tabular}
}
\end{center}
\caption{\label{datasets_statistics} Statistics of our benchmark and OpenIE2016 benchmark. }
\end{table} 

\noindent
{\bf Comparison Systems:} We compare \system{} against state-of-the-art sequence-labeling systems, OpenIE6~\cite{openie6} and Multi2OIE \cite{multi2OIE}, and sequence-generation system, IMoJIE~\cite{imojie}). We also compare it against traditional non-neural systems designed for extracting compact facts: NestIE~\cite{nestie} and MinIE~\cite{minie}.

\begin{table}
\begin{center}
\resizebox{0.8\linewidth}{!}{
\begin{tabular}{llllll} 
\hline
\textbf{Dataset} & \multicolumn{2}{c}{\textbf{Wire57}} && \multicolumn{2}{c}{\textbf{CaRB}} \\
\cline{2-3} \cline{5-6}
 & Proc & Orig & & Proc & Orig \\
\hline
\# Sentences & 56 & 57 && 577 & 641 \\
\# Triples & 309 & 325 && 2101 & 2715 \\
\hline
\end{tabular}
}
\end{center}
\caption{\label{eval_datasets_statistics} Statistics of evaluation datasets, Wire57 and CaRB, before (Orig) and after processing (Proc). }
\end{table}

\noindent
{\bf Evaluation Datasets and Metrics:} 
{We evaluate the OpenIE systems both automatically and manually on standardized benchmarks. For automatic evaluation, we first assess all systems with CaRB\footnote{\url{https://github.com/dair-iitd/CaRB}} test and Wire57 \footnote{\url{https://github.com/rali-udem/WiRe57}} datasets. Since these datasets are not targeted for compact triples, for a fair comparison we exclude triples that have at least one clause within a constituent.  
Table~\ref{eval_datasets_statistics} shows the statistics of the original and processed datasets. Each dataset also provides its own scoring function. We report precision (P), recall (R), and F1 computed by these scoring functions. Wire57 contains more fine-grained extractions than the CaRB dataset and its scoring function is more rigorous for compact facts since it penalizes over-specific extractions. However, both CaRB and Wire57 scoring functions are based on token-level matching of system extractions against ground truth facts. Moreover, these benchmarks are incomplete, meaning that the gold extractions do not include all acceptable surface realizations of the same fact. These drawbacks encouraged us to additionally perform a fact-centered evaluation using the BenchIE \cite{benchIE} benchmark and scoring paradigm. Finally, we carry out a manual evaluation on 100 sentences to avoid bias towards different scorers.}

\noindent
\textbf{{Implementation Details:}} 
{Since the schema design of the table filling model does not support conjunctions inside constituents, we follow previous work \citep{openie6} and pre-process the sentences into smaller conjunction-free sentences before passing them to the system. }

For a fair comparison to previous work, we use {\it bert-based-uncased} \cite{DBLP:journals/corr/abs-1810-04805} as the text encoder for both the constituent extraction model and constituent linking model. Each model contains nearly 110M parameters. {For both models, we set the max sequence length to 512, initial learning rate to 5e-5, weight decay to 1e-5, and the batch size to 32}. We use AdamW optimizer to fine-tune each model. The batch size is 300 for constituent extraction model and 20 for the constituent linking model, both equipped with early stopping.
We use NVIDIA GeForce RTX 2080 Ti GPU to train both models for a cumulative time of 8 hours.

\section{Experimental Results}
\label{experimental_results}

\begin{table*}[t]
\begin{center}
\resizebox{0.9\linewidth}{!} {
\begin{tabular}{llllllllllllll} 
\hline
\textbf{System} & \multicolumn{6}{c}{\textbf{Wire57}} && \multicolumn{6}{c}{\textbf{CaRB}}\\
\cline{2-7} \cline{9-14}
 & P & R & F1 & ACL & NCC & RPA & & P & R & F1 & ACL & NCC & RPA\\
\hline
NestIE & 35.0 & 15.0 & 21.0 & {\bf 4.65} & 0.07 & 1.16 & & {\bf 53.4} & 32.8 & 40.6 & \textbf{4.29} & 0.08 & 1.21\\ 
MinIE & 31.3 & 30.7 & 31.0 & 4.93 & 0.2 & \textbf{1.6} & & 35.3 & {\bf 50.5} & 41.6 & 4.97 & 0.4 & \textbf{1.57}\\
\hline
IMoJIE & 41.2 & 20.1 & 27.0 & 6.23 & 0.26 & 1.07 & & 48.5 & 44.6 & \textbf{46.5} & 6.43 & 0.39 & 1.08\\ 
OpenIE6 & 27.7 & 19.4 & 22.8 & 5.98 & 0.66 & 1.14 & & 44.3 & 44.5 & 44.4 & 6.26 & 0.56 & 1.29\\ 
Multi2OIE & 33.4 & 18.9 & 24.1 & 5.54 & 0.42 & 1.05 & & 48.2 & 44.5 & 46.3 & 6.06 & 0.42 & 1.08\\
\hline
\system{} & \textbf{41.4} & \textbf{25.8} & \textbf{31.8} & {5.23} & \textbf{0.05} & \textbf{1.37} & & 51.3 & 39.9 & 45.0 & 5.08 & \textbf{0.07} & \textbf{1.32}\\

\hline
\end{tabular}
}
\end{center}
\caption{\label{main_results} Performance of OpenIE systems on Wire57 and CaRB datasets. The three analytic metrics (ACL, NCC, RPA) are discussed in Section \ref{sec:analysis}.}
\end{table*}

\subsection{Automatic Token-level Evaluation}
\label{sec:AutomaticEvaluation}
Table~\ref{main_results} summarizes the performance of OpenIE systems across the CaRB and Wire57 datasets and scoring functions. On the fine-grained Wire57 dataset with a strict Wire57 scorer, \system{} outperforms neural OpenIE systems (by 7.2 - 9 F1 pts) and non-neural systems (by 5.8 - 10.8 F1 pts). 

On the more coarse-grained CaRB dataset, almost all OpenIE systems achieve comparable performance in terms of overall F1 using the CaRB scoring function. The neural systems still outperform non-neural systems in terms of F1, which is in line with previous studies. However, neural OpenIE systems are tuned based on the CaRB scoring function and thus tend to produce extractions that are biased towards this scoring method. Previous works~\cite{openie6} also report issues with the scoring function not being able to handle conjunctions properly. Table \ref{tab:carb_scoring_analysis} shows the limitations of the CaRB benchmark and scoring function through an example. As illustrated, the set of extractions produced by \system{} is more exhaustive than IMoJIE and ground truth extractions. However, the CaRB scoring function assigns an F1 score of \underline{62.0} to IMoJIE extractions, and  \underline{39.7} to \system{} extractions. To resolve incompleteness of the CaRB benchmark and potential bias towards its scoring function, we undertake a fact-centered evaluation, detailed in Section \ref{sec:fact-centric}, and a manual evaluation, described in Section \ref{manual_evaluation}.

\subsection{Fact-centric Evaluation}
\label{sec:fact-centric}
{\cite{benchIE} claims that CaRB and Wire57 benchmarks and scoring functions overestimate a system's ability to extract correct facts. They propose an alternative benchmark and evaluation framework, BenchIE, that exhaustively lists all fact-equivalent extractions and clusters them into fact synsets. The scoring function considers an extraction as correct, if and only if it exactly matches any of the gold extractions from any of the fact synsets. They report Precision, Recall, and F1 based on exact triple matching. 

Table \ref{benchIE_results} shows the performance of different OpenIE systems on BenchIE. As shown, \system{} outperforms all other systems except MinIE. We found that MinIE aims to exhaustively produce different representations of the same fact. In contrast, \system{} follows the setup of neural OpenIE systems and encourages at most one representation per fact. As a result, MinIE produces 1.36x more extractions than \system{}, achieving much higher recall than its neural counterparts. 
}

\subsection{Manual Evaluation}
\label{manual_evaluation}

\begin{table}
\begin{center}
\resizebox{0.85\linewidth}{!} {
\begin{tabular}{lll} 
\hline
\textbf{System} & \textbf{Precision} & \textbf{Compactness}\\
\hline
NestIE & 49.1 (84/171) & 98.8 (83/84)\\ 
MinIE & 58.0 (217/374) & 78.8 (171/217) \\
\hline
IMoJIE & 90.0 (156/173) & 53.2 (83/156) \\ 
OpenIE6 & 78.0 (210/269) & 65.2 (137/210)\\
Multi2OIE & 78.6 (151/192) & 59.6 (90/151)\\
\hline
\system{} & 75.8 (175/231) & 94.9 (166/175) \\
\hline
\end{tabular}
}
\end{center}
\caption{\label{minimality_precision} Manual evaluation of OpenIE systems on CaRB validation set. Precision indicates the percentage of correct extractions. Compactness indicates the percentage of compact extractions amongst the correct ones.
}
\end{table}

{Limitations in the aforementioned benchmarks and evaluation frameworks encouraged us to perform human evaluation on triples generated by various systems. To this end, we randomly select 100 sentences from the CaRB validation set and feed them to all systems to investigate the generated triples.} 
Next, we ask two graduate CS students, blind to the OpenIE systems, to mark each triple for correctness (0 or 1) based on whether it is asserted in the text and correctly captures the semantic information. They also label extractions for compactness (0 or 1). We consider an extraction compact if  none of its constituents is longer than 10 words, includes conjunction or can be an independent extraction. We found an inter-annotator agreement of 0.68 on correctness and 0.83 on compactness using the Cohens Kappa metric. We report the results of the manual evaluation in Table~\ref{minimality_precision}. Neural systems target informativeness, which results in high precision at the cost of compactness. On the other hand, non-neural systems that aim for compact triples suffer from low precision. \system{} offers a better trade-off between precision and compactness. It achieves comparable precision to neural models (-6 \%) while providing substantially more compact extractions (+36 \%). Compared to the MinIE, \system{} produces triples with significantly higher precision (+22 \%) while producing a comparable number of compact triples. NestIE achieves comparable compactness rate to \system{} but suffers from low precision and total number of extractions.

\begin{table}[t]
\begin{center}
\resizebox{0.7\linewidth}{!}{
\begin{tabular}{llll} 
\hline
\textbf{System} & \multicolumn{3}{c}{\textbf{BenchIE}}\\
\cline{2-4}
 & P & R & F1  \\
\hline
NestIE & 37.1 & 10.2 & 16.0\\ 
MinIE & \textbf{42.9} & \textbf{27.8} & \textbf{33.7}\\
\hline
IMoJIE & 34.3 & 12.8 & 18.6\\ 
OpenIE6 & 31.1 & \textbf{21.4} & 25.3\\ 
Multi2OIE & 39.2 & 16.1 & 22.8\\
\hline
\system{} & \textbf{40.3} & 19.0 & \textbf{26.2}\\

\hline
\end{tabular}
}
\end{center}
\caption{\label{benchIE_results} Performances of OpenIE systems on the BenchIE dataset.}
\end{table}

\section{Analysis}
\label{sec:analysis}
\subsection{Compact and Overlapping Constituents}
To understand the performance of \system{} in generating compact triples that share constituents, we introduce the following metrics:

\begin{itemize}[noitemsep,topsep=0pt,leftmargin=*]
\item Average Constituent Length (ACL): average length of constituents across all system-generated triples. This is a ``syntactic" measure of compactness. The lower the ACL score, the higher the compactness of triples.

\item Number of Constituent Clauses (NCC): average number of clauses per constituent that could be extracted as independent triples. The lower the NCC score, the better the compactness of triples.

\item Repetitions Per Argument (RPA): number of total arguments divided by the number of unique arguments. The higher the RPA score, the higher fraction of total constituents produced per sentence are shared.

\end{itemize}

Table \ref{main_results} summarizes the performance on these metrics over CaRB and Wire57 benchmarks. We do not conduct a separate analysis over BenchIE since it uses a subset of CaRB sentences. As shown, the ACL scores of \system{} are significantly lower than its neural counterparts and closely follows MinIE. The average constituent length (ACL) of NestIE triples is the lowest since it breaks sentences into small triples with verb, noun, preposition, and adjective mediated relations. For instance, the sentence: ``2 million people died of AIDS.'' is broken down into T1: (2 million people; died), and T2: (T1; of; AIDS). However, its fine-grained strategy greatly sacrifices F1 for compactness.  
\system{} achieves the lowest NCC score which indicates that the constituents in triples contain the fewest verbal clauses. As a result, these triples are more suitable for downstream applications such as text summarization and knowledge-base construction than other counterparts.

Finally, high RPA scores of \system{} demonstrate the effectiveness of our approach as it enables the system to reuse the same constituent to generate multiple triples. MinIE achieves a slightly higher RPA score than \system{} since it extracts multiple triples to represent the same fact leading to a higher repetition of unique constituents.

\subsection{Effectiveness of Design Choices}
\noindent
{\textbf{Pipelined Approach vs. Unified Table Filling}.}
To compare our pipelined approach with a unified extraction model, we follow UniRE ~\cite{wang2021unire}, which decodes a single table to identify entities and relations jointly. We follow their 3-step decoding algorithm to obtain the constituents and links between them from the same table (with the schema shown in Figure~\ref{fig:constitunet_table}). We refer to this model as \system{}\textsubscript{uni}. We report the performances in  Table~\ref{joint_systems_comparison} and show that performance drops by jointly training the constituent and linking model. This aligns with the observations in recent entity-relation extraction work that pipelined approaches are more effective than joint models.

\noindent
{\textbf{Effectiveness of Schema Design}.} Our table schema for constituent extraction includes both labels for constituents as well as labels to link them. We argued that this design captures the contextual dependency information between the constituents that boosts extraction performance. We compare the effectiveness of this schema design to another schema that uses only constituent labels $Y_c: \{Argument, Predicate\} \cup None$. Note that we use the same constituent linking model to obtain triples from the extracted constituents. We refer to this setting as \system{}\textsubscript{const table}. Table~\ref{joint_systems_comparison} illustrates the performance of this system on both CaRB and Wire57 datasets. We find that \system{} achieves significantly higher F1 compared to \system{}\textsubscript{const table} and conclude that incorporating additional context in the table schema improves the performance of the constituent extraction model.

\noindent

\begin{table}
\begin{center}
\resizebox{0.7\linewidth}{!}{
\begin{tabular}{ lll } 
\hline
Method & \textbf{Wire57} & \textbf{CaRB}\\
\hline
\system{} & 31.8 & {  45.0} \\
\system{}\textsubscript{uni} & 17.6 & { 35.8} \\ 
\system{}\textsubscript{const table} & 26.0 & { 40.1} \\
\hline
\end{tabular}
}
\end{center}
\caption{\label{joint_systems_comparison} Comparing F1 scores of CompactIE against joint extraction systems.}
\end{table}

\subsection{{Error Analysis}}

\begin{table*}[t]
\begin{center}
\resizebox{0.8\linewidth}{!}{
\begin{tabular}{lllll} 
\hline
\textbf{System} & Subject & Predicate & Object & \textbf{F1} \\
\hline

\multirow{3}{*}{\textbf{\small Gold}} & {Applications} &  {use this service to record} & {activity for a system} & \\
& {other OSIDs} & {use the service to record} & {data} & - \\
& {other OSIDs} & {use the service to record data} & {during analysis} & \\
\hline
\multirow{2}{*}{\textbf{\small IMoJIE}} & {Applications} &  {use this service to record} & {activity for a system} & \multirow{2}{*}{62.0} \\
& {other OSIDs} & {use} & {the service to record data during ... analysis} & \\
\hline
\multirow{5}{4em}{\textbf{\small \system{}}} & {Applications} &  {use} & {this service to record activity for a system} & \multirow{5}{*}{39.7} \\
& {other OSIDs} & {use} & {service to record data during development} \\
& {other OSIDs} & {use} & {the service} \\
& {the service} & {record} & {data during debugging} \\
& {the service } & {record} & {data during analysis}\\
\hline
\end{tabular}
}
\end{center}
\caption{\label{tab:carb_scoring_analysis} Gold, IMoJIE and \system{} extractions for the sentence: ``{\it Applications use this service to record activity for a system while other OSIDs use the service to record data during development, debugging, or analysis.}'' and their CaRB F1 score that evaluates extractors triples against gold triples.}
\vspace{-1em}
\end{table*}

We examine \system{} triples produced for 50 randomly selected sentences of the CaRB validation dataset and 20 randomly selected sentences of the Wire57 dataset. Upon close analysis, we identify five major sources of error:

\noindent
\textbf{Constituent Not Found}: (49.29\% ) We find that the constituent extraction model can fail to correctly label the constituents in the table. We found that the model gets biased towards producing $None$ labels due to the imbalanced distribution of labels. 

\noindent
\textbf{Wrong Relation Type}: (28.17\%) These involve errors where the constituent linking model fails to correctly predict the link between the constituents. The current model encodes one sentence per predicate to find its arguments. Alternatively, we can encode one sentence per predicate-argument pair to focus more on each relation. Relation labels in the constituent extraction model can also assist the linking model in predicting the correct relations. We reserve this issue for future work.

\noindent
\textbf{Boundary Detection Error}: (11.26\%) These include errors where the decoder in constituent extraction fails to correctly identify the boundaries of the constituents. Boundary detection in constituent extraction model is highly dependant on the choice of distance threshold ($\alpha$), as explained in \ref{sec:2stepDecoding}, which limits its robustness. 

\noindent
\textbf{Inexpressive Table Error}: (7.04\%) These include errors where constituents have overlapping spans that participate in two roles within the same extraction or two different extractions.

Less than 4.22\% of the errors were because of incorrect constituent type predictions. This indicates the effectiveness of our table filling method on constituent type detection.

\section{Related Work}

OpenIE has been studied extensively for over a decade with a history of statistical and rule-based systems~\cite{textRunner,fader2011identifying,clausie,schmitz2012open,angeli2015leveraging} that extract triples from sentences without using any training data. Recently, neural models have been developed that are trained end-to-end on extractions bootstrapped from previous OpenIE systems. These can broadly be classified into {\it labeling-based} and {\it generation-based} systems. 

Labeling-based systems~\cite{stanovsky2018supervised,openie6,multi2OIE} tag each word in the sentence and construct triples in an auto-regressive manner or by using a unique predicate for each triple. Generation-based  systems~\cite{imojie,bhutani2019open} use a sequence-to-sequence model to generate triples one word at a time. Labeling-based systems can handle redundancy in extracted triples and are faster than generation-based systems~\cite{openie6}.

\noindent
{\bf Compactness in OpenIE}: There has been prior work~\cite{nestie,minie,stanovsky-2016,angeli2015leveraging} that focuses on finding compact triples and shows that concise triples are useful in several semantic tasks. However, recent studies~\cite{lechelle2018wire57,gashteovski2020aligning} indicate that neural OpenIE systems produce more specific triples with additional information than conventional OpenIE systems and are harder to align with existing knowledge bases. Therefore, we focus on designing a neural OpenIE system that extracts compact triples. 

\noindent
{\bf Grid Labeling}: Also known as table filling, grid labeling has been recently applied to entity relation extraction~\cite{gupta2016table,wang2021unire} and open information extraction tasks~\cite{imojie}. However, these models map entities (constituents) and relations (subject, object) in a unified label space to capture the inter-dependency between them. \citep{DBLP:journals/corr/abs-2010-12812} shows that a pipelined approach for entity and relation extraction outperforms prior joint models that use the same encoder for the two sub-tasks. In this work, we validate this claim for the OpenIE task. Furthermore, we design a grid labeling schema that identifies constituents and their types, akin to entities in the entity relation extraction task.

\section{Conclusion}
\label{sec:conc}
In this work we extract compact triples from single sentences using an end-to-end pipelined approach, first extracting triple constituents using a novel table filling model and then determining relations between them with a classifier. Our method achieves excellent performance in producing exhaustive compact triples with high precision. We hope that \system{} serves as a strong baseline and makes us re-think the value of all-at-once information extraction systems.  

\section{Acknowledgments}
\label{sec:ack}
The research described herein was sponsored by the U.S. Army Research Institute for the Behavioral and Social Sciences, Department of the Army (Contract No. W911NF-20-C-0028). The views expressed in this presentation are those of the author and do not reflect the official policy or position of the Department of the Army, DOD, or the U.S. Government.

\bibliography{acl2022}
\bibliographystyle{acl_natbib}

\appendix
\section{Appendix}
\label{sec:appendix}

\subsection{Benchmark Creation}
\label{sec:benchmarkCreation}

The Algorithm \ref{algo:ext} gives a high-level overview of our benchmark creation mechanism while a lot of details and difficulties have been omitted. The Benchmark Creation Algorithm extracts triples for each sentence using the Algorithm \ref{algo:exttriples}. The OpenIE system used to produce triples out of simple clauses is IMoJIE \cite{imojie}.

\begin{figure}[t]
\centering
\includegraphics[width=0.45\textwidth]{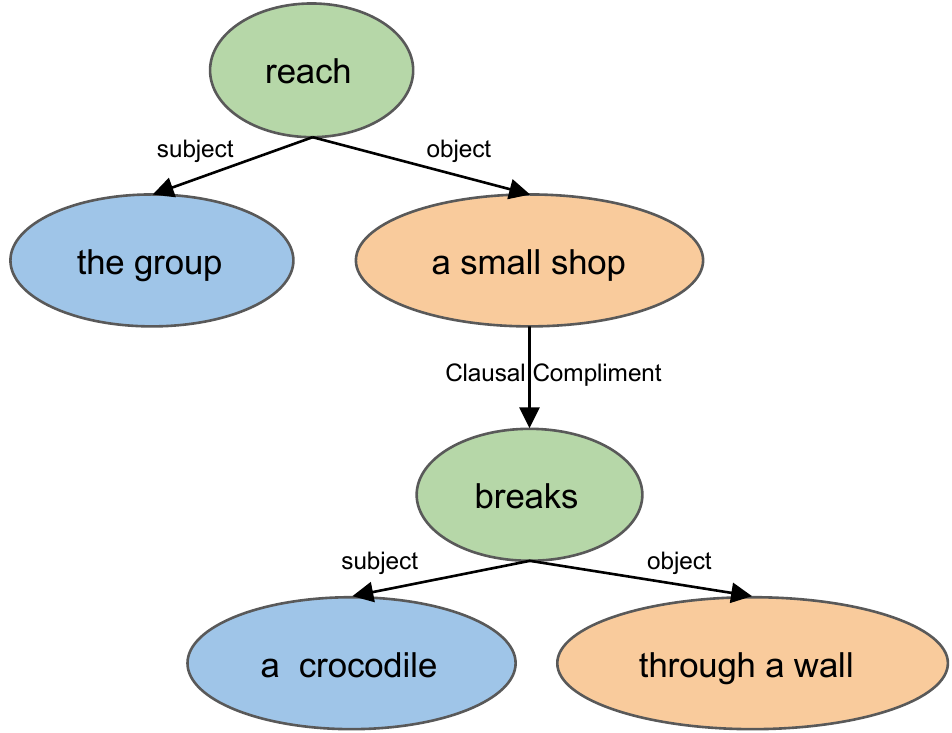}
\caption{Sentence Tree for input sentence: \emph{``The group reach a small shop, where the crocodile breaks through a wall"}.}
\label{fig:BenchmarkExample}
\vspace{-1em}
\end{figure}

The following example illustrates the benchmark creation algorithm. Given the sentence: \emph{``The group reach a small shop, where the crocodile breaks through a wall"}, the algorithm first builds the sentence tree as shown in Figure \ref{fig:BenchmarkExample}. 
Then, starting from the root, \emph{ExtractTriple} function traverses the tree until it reaches a child (``breaks") with no further clausal children. At this point, a clause for the subtree rooted at ``breaks" is generated and fed into the IMoJIE system. IMoJIE extracts triple: (the crocodile; breaks; through a wall) out of this clause. Then, since both children of the root (``reach") are processed, the IMoJIE triple of the root's corresponding clause is extracted as (The rest of the group; reach; a small shop).

\begin{algorithm}
    \KwData{Tree Node R}
    \KwResult{Set of compact triples T}
    T = set() \;
    \For{child in R.children}{
        \If{child has no clausal child}{
            T += IMoJIE(child.clause) \;
        }
        \Else{
            T+= ExtractTriples(child) \;
        }
    }
    T += IMoJIE(R.clause) \;
    \Return T                
    \caption{ExtractTriples}
    \label{algo:exttriples}
\end{algorithm}

\begin{algorithm}
    \KwData{Sentence List $S = [s_1, s_2, .., s_n]$}
    \KwResult{B benchmark of compact triples for sentences in S}
    B = set() \;
    \For{sentence in S}{
        root = build\_sentence\_tree(sentence) \;
        B += ExtractTriples(root) \;
        
    }
    \Return B
    \caption{Benchmark Creation}
    \label{algo:ext}
\end{algorithm}

\subsection{Table Decoding}
\label{sec:2stepDecoding}
Following the \cite{wang2021unire} work, in the testing phase, we rely on the label probability tensor $\mathcal{P} \in \mathbb{R}^{|s|\times |s| \times |Y|}$ of the sentence $s$, to first extract constituent spans, and then predict the constituent type. Next, we describe the decoding procedure.

\subsubsection{Constituent Span Detection}
One important observation of the ground truth table is that a constituent's corresponding rows and columns are identical (e.g., row 2 and row 3 of Figure \ref{fig:constitunet_table} are identical). Therefore, given the tensor $\mathcal{P}$, we compute the distance of adjacent rows (and columns). If the distance is larger than a pre-defined threshold $\alpha$ (which is set to 1.2), a split position is detected. This means that the two adjacent rows (columns) belong to different constituents or one belongs to a constituent while the other is not. Following the \cite{wang2021unire} work, we flatten the $\mathcal{P}$ tensor from both row and column perspectives and calculate the euclidean distance of adjacent rows and adjacent columns. Finally, we average these two distances as the final distance and compare the final  distance to $\alpha$ to find the span of different constituents.
\subsubsection{Constituent Type Detection}
Given a constituent's span $(i, j)$, we decode the constituent type $t^* \in Y$, where $Y = Y_c \cup Y_r \cup None$, according to its corresponding square symmetric about the diagonal:
\[t^* = argmax_{t \in \{Y_c \cup None\}}Avg(P_{i:j,i:j,t})\]
Spans with predicted type $t^* \in Y_c$ are regarded as constituents and passed to the linking model.

\end{document}